\documentclass[11pt,a4paper]{article}
\pdfoutput=1
\usepackage[hyperref]{acl2021}
\usepackage{times}
\usepackage{latexsym}
\usepackage{graphicx}
\usepackage{pgfplots}
\usepackage{arydshln}
\pgfplotsset{compat=1.14}

\usepackage{microtype}

\aclfinalcopy %

\newcommand{\err}[1]{{\color{red} \underline{#1}}}
\newcommand{\scare}[1]{`{#1}'}

\definecolor{purple}{HTML}{332288}
\definecolor{green}{HTML}{117733}
\definecolor{teal}{HTML}{44AA99}
\definecolor{lightblue}{HTML}{88CCEE}
\definecolor{beige}{HTML}{DDCC77}
\definecolor{salmon}{HTML}{CC6677}
\definecolor{pink}{HTML}{AA4499}
\definecolor{maroon}{HTML}{882255}

\newcommand{\errname}[1]{{\color{green} \underline{\textbf{#1}}}}
\newcommand{\errnumber}[1]{{\color{purple} \underline{\textbf{#1}}}}
\newcommand{\errword}[1]{{\color{salmon} \underline{\textbf{#1}}}}

\newcommand{\lineacross}{\rule{\linewidth}{1pt}}

\title{Generation Challenges: Results of the Accuracy Evaluation Shared Task}

\author{Craig Thomson \\
  Dept of Computing Science \\
  University of Aberdeen \\
  Aberdeen, UK\\
  \texttt{c.thomson@abdn.ac.uk} \\\And
  Ehud Reiter \\
  Dept of Computing Science \\
  University of Aberdeen \\
  Aberdeen, UK\\
  \texttt{e.reiter@abdn.ac.uk} \\}

\date{}

\begin{document}
\maketitle
\begin{abstract}
The Shared Task on Evaluating Accuracy focused on techniques (both manual and automatic) for evaluating the factual accuracy of texts produced by neural NLG systems, in a sports-reporting domain.
Four teams submitted evaluation techniques for this task, using very different approaches and techniques.  The best-performing submissions did encouragingly well at this difficult task.
However, all automatic submissions struggled to detect factual errors which are semantically or pragmatically complex (for example, based on incorrect computation or inference).
\end{abstract}

\section{Introduction}

Users expect data-to-text natural language generation (NLG) systems to generate textual summaries which are accurate.  However, many NLG systems, especially neural ones, generate texts which are factually incorrect.

The most reliable way to assess the accuracy of a generated text is to ask human annotators to carefully fact-check the text.  However this is a time-consuming and expensive process.  In earlier work, we developed a protocol \citep{INLG20accuracy} where three Mechanical Turk workers (who had been screened and passed a qualifying test) carefully annotated factual errors in a text produced by a neural NLG system.  The protocol was effective and showed high interannotator agreement, but it took annotators 20-30 minutes (each) to fact-check a moderately complex 300-word paragraph produced by a neural data-to-text NLG system.   The total cost of the process (including fees to Amazon and money spent on the screening process for potential annotators) was about US\$30 per text.

It would be very useful to the NLG community if we could come up with quicker and easier ways of measuring accuracy and factual correctness which have good correlations with the protocol of \citet{INLG20accuracy}.   Such methods could be based on less time-consuming human evaluations or on automatic metrics.   However, these techniques should only be used if they have good agreement and correlation with careful high-quality human fact-checking by multiple annotators.

In this shared task, participating teams submitted techniques (both human and automatic) for evaluating the factual accuracy of summaries of basketball games produced from box score (and other game data) by three neural NLG systems.  These techniques were evaluated by computing precision and recall (of identified factual errors) against a gold-standard human annotation produced by \citet{INLG20accuracy}'s protocol.  Some of the systems did well overall, but it was also clear that some types of factual errors are difficult to detect.

We hope that our shared task encourages researchers from many fields to work on the problem of identifying factual errors in generated texts;  progress in this area would be very helpful for NLG.  Full details of the shared task requirements, as well as both the training and test corpus can be found at \href{https://github.com/ehudreiter/accuracySharedTask}{https://github.com/ehudreiter/accuracySharedTask}.

\section{Task}
\label{sec:task}

Participants were asked to submit a technique for identifying incorrect statements in a generated text.   This meant statements which are not true in the real world (ie, classic fact-checking), not just statements which disagree with (or are not derivable from) the system run-time data (see Section 3.1 of \citet{INLG20accuracy}).  Techniques could be
\begin{itemize}
    \item Human evaluation protocols.  Subjects would have access to data about the game and the teams, and also (if part of the protocol) to a human-authored reference text.
    \item Automatic metric (algorithm).  The algorithm will have access to data about the game and the teams, and to a reference text.
    \item A combination of human evaluation and automatic metrics.
\end{itemize}

The output of the evaluation protocol or metric was a list of mistakes in the text.  Each mistake was characterised by
\begin{itemize}
    \item Its position in the text (start token and end token).
    \item A category.  We use the following categories, which are based on \citet{INLG20accuracy}
    \begin{itemize}
        \item {\em Incorrect number:}  It does not matter whether the number is spelled out or is in digits.
        \item {\em Incorrect name (for named entities):} In a sports reporting context, this includes people, places, teams, and days of the week.
        \item {\em Incorrect word:} A word which is not one of the above and is incorrect.
        \item {\em Context error:} A phrase which causes an incorrect inference because of context or discourse.
        \item{\em Not checkable:} A statement which can not be checked, either because the information is not available or because it is too time-consuming to check.
        \item {\em Other:}  Any other type of mistake.
    \end{itemize}
\end{itemize}
An example is shown in \autoref{fig:example}.  Note that this example combines fragments from texts produced by several different systems, along with some manual adjustments, in order to illustrate different types of mistakes in a simple way.

\begin{figure}[!b]
\lineacross{}
The Memphis Grizzlies (5-\err{2}) defeated the Phoenix Suns (3 - 2) \err{Monday} 102-91 at the \err{Talking Stick Resort Arena} in Phoenix. The Grizzlies had a \err{strong} first half where they \err{out-scored} the Suns \err{59}-\err{42}. Marc Gasol scored 18 points, \err{leading} the Grizzlies.  \err{Isaiah Thomas added} 15 points, he is \err{averaging 19 points on the season so far}.

\vspace{5mm}
List of errors:
\begin{itemize}
    \item \err{2}: incorrect number, should be 0.
    \item \err{Monday}: incorrect named entity, should be Wednesday.
    \item \err{Talking Stick Resort Arena}: incorrect named entity, should be US Airways Center.
    \item \err{strong}: incorrect word, the Grizzlies did not do well in the first half.
    \item \err{out-scored}: incorrect word, the Suns had a higher score in first half.
    \item \err{59}: incorrect number, should be 46.
    \item \err{42}: incorrect number, should be 52 .
    \item \err{leading}: incorrect word.  Marc Gasol did not lead the Grizzlies, Mike Conley did with 24 points.
    \item \err{Isaiah Thomas added}: context error.  Thomas played for the Suns, but context here implies he played for the Grizzlies and added to their score.
    \item \err{averaging 10 points on the season so far}: not checkable.  This is very hard to check, since data sources report performance per season and per game, not performance up to a particular point in a season.
\end{itemize}
\caption{Example text with error annotations.  Corrections and explanations are not required, but are included here for clarity. Box score data for this game is available at	\url{https://www.basketball-reference.com/boxscores/201411050PHO.html} .}
\label{fig:example}
\lineacross{}
\end{figure}

\section{Data}
\label{sec:data}

We manually annotated, using the procedure of \citet{INLG20accuracy}, 90 texts produced by three neural NLG systems that use basketball box score data: \citet{wiseman-etal-2017-challenges}, \citet{puduppully-2019-planning}, and \citet{10.1007/978-3-030-45439-5_5}.  In total, 30 texts were annotated from each system.  Of these, 60 texts (20 from each system) were given to shared task participants as training data, and 30 texts (10 from each system) were reserved for a separate test set, which participants did not see until they had submitted their solutions.

Annotators were recruited on the Amazon Mechanical Turk platform.  Fair treatment and compensation of workers is essential \citep{Silberman-responsible-2018}, not only from an ethical standpoint, but to ensure high quality annotations.  We paid annotators approximately US\$20 per hour.  The same three annotators marked up all 90 texts.

\subsection{Systems Used}

The three neural systems we used
explored different ways of modifying the neural architecture.  The system of \citet{wiseman-etal-2017-challenges} defined the Rotowire task and provided initial benchmarks for machine translation systems using copy attention, it is included for this reason.  \citet{puduppully-2019-planning} learned a document plan which was then used to generate text, whilst \citet{10.1007/978-3-030-45439-5_5} used a hierarchical encoder to group attributes (such as statistics) by their respective entities (players/teams).

Other systems in this domain which could be used for evaluation include \citet{puduppully-2019-entity}, \citet{wang-2019-revisiting}, \citet{gong-etal-2019-table}, and \citet{iso-etal-2019-learning}.  Our aim, however, is to assess how well results produced by the participant's evaluation techniques correlate with the gold-standard fact-checking.  Hence we are looking for a set of systems which generate texts that contain a significant number of accuracy errors, not complete coverage of all systems that generate texts from basketball box score data. 

\subsection{Multiple Correct Annotations}
\label{sec:complex_annotations}
Sometimes there are multiple correct ways of annotating errors.  For example, consider the sentence
\begin{quote}
Lou Williams led the team in scoring, dropping 30 points, six rebounds and seven assists
\end{quote}{}
Suppose that it was another player, Solomon Hill, who had 30 points, 6 rebounds, and 7 assists.  In this case, the sentence could be corrected either by changing the player name (to Solomon Hill), or by changing the statistics (to the correct ones for Lou Williams).  In such cases we asked annotators to try to find the smallest number of annotations required to correct the sentence, prioritising categories in the order of Name, Number, Word, Context, Other, Not checkable.  This is straightforward in this example. where the choice is correcting a single player name, or three numbers.

There were, however, a few cases where multiple complex annotations were plausible and the preferred one was not clear to our annotators.   For example, in our test we encountered a sentence that was marked up by annotators as shown in \autoref{fig:complex}:

\begin{figure}[ht]
    \begin{quote}
        \textbf{Annotator T1}: The \errword{only other} \errname{Raptor} to reach double figures in points was \errname{Dwyane} Dragic, who \errword{came off the bench} for 22 points (\errnumber{9}-\errnumber{17} FG, 3-7 3Pt, 3-3 FT), \errnumber{six} rebounds and five assists.
        
        \textbf{Annotator T2}: The \errword{only other} \errname{Raptor} to reach double figures in points was \errname{Dwyane Dragic}, who came \errword{off the bench} for 22 points (\errnumber{9}-\errnumber{17} FG, 3-7 3Pt, 3-3 FT), \errnumber{six} rebounds and five assists.
        
        \textbf{Annotator T3}: The \errword{only other} Raptor to reach double figures in points was \errname{Dwyane Dragic}, who came off the bench for \errnumber{22} points (9-\errnumber{17} FG, \errnumber{3}-7 3Pt, \errnumber{3}-\errnumber{3} FT), \errnumber{six} rebounds and \errnumber{five} assists.
    \end{quote}
    
    \caption{Annotations by each annotator, showing \errname{Name}, \errnumber{Number}, and \errword{Word} errors.}
    
    \label{fig:complex}

\end{figure}

T1 and T2 essentially decided to change the player name to \textit{Goran Dragic}; since \textit{Dragic} played for the other team (\textit{Heat}), they also corrected \textit{Raptors}.  They then corrected three of the numbers accordingly and noted that \textit{Dragic} did not come off the bench, he started the game.  T3 disagreed, changing the player name to \textit{Lou Williams} who did in fact start for the \textit{Raptors}.  Whilst this minimised Name and Word errors, it required correcting 7 of the numbers, leading to 9 errors in all, compared to the 7 errors annotated by T1 and T2.

The majority annotation (T1 and T2) was correct in this case according to our \scare{choose annotation with smallest number of errors}.  But it is not trivial for annotators to search through multiple possible annotations looking for the optimal one, and in a larger sense it is not clear which annotation is \scare{correct}.

\section{Accuracy Errors Observed}
\label{sec:accuracy_errors}
\begin{table*}[!ht]
    \centering
    \resizebox{\textwidth}{!}{%
    \begin{tabular}{l|l| r | l }
         Error & Type & count & note \\ \hline
         NUM-DIGIT & Number & 270 & number in digits, such as an incorrect quantity of points \\
         TEAM & Name & 162 & name of team, such as \emph{Miami Heat} \\
         NUM-WORD & Number & 130 & a number spelled as a word or words \\
         DAY-WEEK & Name & 128 & a day of the week, such as \emph{Wednesday} \\
         PLAYER & Context & 50 & player name (used in incorrect context) \\
         led & Word & 40 & word \emph{led}, often indicates a player led their team by some measure \\
         a (an) & Number & 34 & \emph{a} or \emph{an} meaning the number 1 \\
         ORDINAL & Number & 26 & ordinal number often describing consecutive games\\
         double-double & Word & 23 & word \emph{double-double}, a \href{https://en.wikipedia.org/wiki/Double-double}{basketball metric} \\
         PLAYER & Name & 21 & name of a player, such as \emph{LeBron James} \\
    \end{tabular}
    }
    \caption{Errors that occurred at least 20 times in the training data. NUM-DIGIT, TEAM, NUM-WORD, DAY-WEEK, ORDINAL refer to types of words.  Number, Name, Context, Word refer to types of errors.}
    \label{tab:errorsobserved}
\end{table*}

In this section we discuss and give some insights about the accuracy errors we observed in the manually-annotated training data (i.e, the 60 annotated texts given to participants as training data).  We look separately at the different types of errors listed in \autoref{sec:task}, and also at the impact of position and the neural NLG system used.
\autoref{tab:errorsobserved} lists all errors that occurred at least 20 times in the training data.

\subsection{Number errors}
Number errors are the most common type of error in our corpus; there were 474 Number errors in the 60 texts in the training data.  This category includes errors in numbers presented as digits (NUM-DIGIT), errors in spell-out numbers (NUM-WORD), and errors when a/an is used to mean the number 1.

From a semantic perspective, we can distinguish between \emph{errors in copying numbers from the data} (eg, claiming that Smith scored 20 points when the box score data says that he scored 10 points) and \emph{errors in calculating numbers which are not directly in the data} (eg, calculating the score at half-time, from the quarter-level scores given in the box office data).  Both types of errors were common in our corpus. 

\subsection{Name errors}
There were 317 Name errors in our corpus.  TEAM, PLAYER, and DAY-WEEK (from \autoref{tab:errorsobserved}) are all examples of a Name error.  Many of these errors arose when NLG systems tried to create sentences for which they lacked data, such as the following:
\begin{quote}
    The Sixers' next game will be at \errword{home} against the \errname{New Orleans Pelicans} on \errname{Wednesday}
\end{quote}

Information about the next game is not present in the data used by the three systems which were fact-checked, so they simply guessed team and day of week, and usually guessed wrong.  Of course we cannot expect a system to generate accurate texts that communicate information which is not present in the input data!  But we can expect data-to-text systems to avoid sentences which communicate unavailable data.

As mentioned in \autoref{sec:complex_annotations}, sometimes a sentence could be characterised as having either a Name or a Number error.
In such cases we asked annotators to make the correction which required the smallest number of changes.

\subsection{Word errors}
\label{sec:accuracy_errors:word_errors}
There were 334 Word errors in our corpus.  They
can be divided into two categories: errors in using words with clear unambiguous definitions (such as \emph{out-scored} in \autoref{fig:example}) and errors in words with fuzzy definitions (such as \emph{strong} in \autoref{fig:example}).

The most common error in a well-defined word is \emph{double-double}. A double-double occurs when a player has ten or more (double-digits) in exactly two of the following categories: points, rebounds, assists, steals, and blocks.  Note that if a player has ten or more in three of the categories, this is called a \emph{triple-double} (3 statistics in double-digits) rather than a \emph{double-double}.  While double-double is easy to define via rules, there were 23 mistakes in our 60 corpus texts (Table~\ref{tab:errorsobserved}) in the usage of this word; this seems to be a difficult concept for our systems to learn.

The most common error in a fuzzy word was \emph{led}.  \emph{Led} appears in many contexts, for example we can say that a player \emph{led the bench in scoring} or that a team \emph{led at the half}. 

The meaning of \emph{led} is not clear-cut, and indeed on a few occasions the annotators disagreed on whether \emph{led} was appropriate. This is because \emph{led} (when used in descriptions of basketball games) can encompass rebounds, assists, steals and blocks as well as points.   For example, if one player has 25 points, 0 assists and 0 rebounds, and a second player has 22 points, 10 assists, and 10 rebounds, then the first player led in scoring, but it could be argued that the second player had a stronger performance overall, and therefore \emph{led}.  However, most of the incorrect usages of \emph{led} marked by the annotators were in cases where all of the annotators agreed that \emph{led} was inappropriate.

Some \emph{ORDINAL} errors were related to this.  For example, one sentence would state that a player \emph{led}, and the subsequent sentence would state that a second player \emph{was second}.

\subsection{Context error}
A Context occur occurs when a statement is literally true but misleading in context.  There were 51 Context errors in our corpus, 50 of which involved PLAYERs.  Typically the text would mislead the reader as to a player's status, especially which team he is playing for.  An example from \autoref{fig:example} is:
\begin{quote}
    Marc Gasol scored 18 points, leading the Grizzlies.  Isaiah Thomas added 15 points
\end{quote}
Thomas scored 15 points but played for the other team (Suns).  This is a Context error, since the context implies that Thomas played for the Grizzlies.

Such errors were common, the systems had a difficult time in learning when it is contextually appropriate to mention a person.

\subsection{Not Checkable and Other}
A Not Checkable error occurs when the annotator cannot check whether a fact is true or not.  There were 37 such errors in our corpus.  They
usually occurred when complex statistics were given which were difficult and time-consuming for annotators to check.  In order to keep the annotation task manageable, annotators were told not to look at more than 4 previous games. This made it impossible to check statements such as \emph{he is averaging 19 points on the season so far} (from \autoref{fig:example}), which requires looking at data from every previous game in the season.

We discouraged our annotators from using the Other category unless absolutely necessary, and in fact there was only one Other error in our corpus, which was the nonsensical statement \emph{run with the first - round win of the season}.

\subsection{Position analysis}
\label{sec:accuracy_errors:position_analysis}

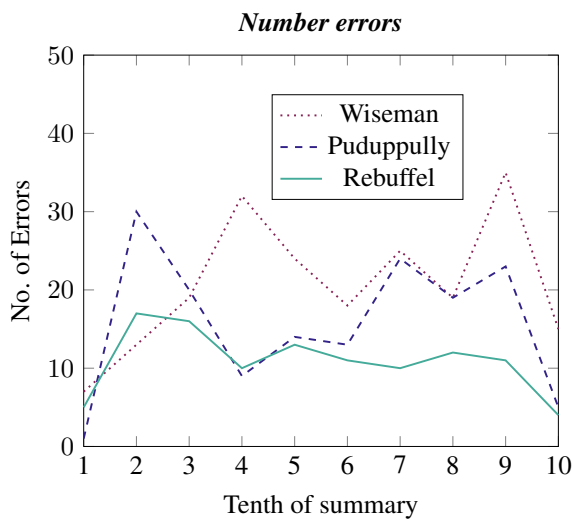
\begin{figure}[ht!]
    \resizebox{\columnwidth}{!}{%
    \begin{tikzpicture}
        \begin{axis}
            [
                ymin=0,
                every axis x label/.style={at={(current axis.right of origin)},anchor=south east},
                title={\textit{\textbf{Number errors}}},
                xlabel=Tenth of summary,
                ylabel=No. of Errors,
                x label style={at={(axis description cs:0.5,-0.1)},anchor=north},
                enlargelimits = false,
                xticklabels from table={number.dat}{X},xtick=data,
                ymax=50,
                legend style={at={(0.8,0.9)}}
            ]
            \addplot[maroon,dotted,thick] table [y=Wiseman,x=X]{number.dat};
            \addlegendentry{Wiseman}
            \addplot[purple,dashed,thick] table [y=Puduppully,x=X]{number.dat};
            \addlegendentry{Puduppully}
            \addplot[teal,thick] table [y= Rebuffel,x=X]{number.dat};
            \addlegendentry{Rebuffel}];
        \end{axis}
    \end{tikzpicture}
    }%
    \caption{Number errors in different tenths of the summary.}
    \label{Figure-Number-errors}
\end{figure}

In addition to analysing errors by category, we also wondered if there might be fewer errors at the beginning of the text, and more at the end.  There was in fact a sharp increase in Name errors in the last sentence (\autoref{Figure-Name-errors}), but this was probably due to the fact that the last sentence usually described next games, and the systems did not have this information so they hallucinated.  Other than this, we did not see an increase in errors later in the text.  \autoref{Figure-Number-errors} shows the distribution of Number errors in different positions, the other error types (excluding Name) have a similar distribution.  For both of these figures, error counts are shown based upon which tenth of the summary (by token id) the error starts in.

\subsection{System analysis}
\label{sec:accuracy_errors:system_analysis}

Last but not least, we wanted to look at the error profiles for the three systems we included in the error corpus.
Two of the systems used RNN-based encoders \citep{wiseman-etal-2017-challenges,puduppully-2019-planning} and the third used a hierarchical transformer \citep{10.1007/978-3-030-45439-5_5}.  \autoref{fig:all_system_errors} shows the errors each system had within each category.  The hierarchical transformer made fewer Number errors than both RNN based systems but more Context errors.  It is unclear why the hierarchical encoder of \cite{10.1007/978-3-030-45439-5_5} made more Context errors, although it may be learning to better group entities with their attributes, at the expense of ordering between entities.

\begin{table*}[ht!]
    \begin{tabular}{lccccccc}
        \hline
        System     & encoder & NAME & NUMBER & WORD & CONTEXT & NOT\_CHECK & OTHER \\ \hline
        Wiseman & RNN & 5.9 & 10.4 & 6.7 & 0.3 & 1.0 & 0.0 \\
        Puduppully & RNN & 5.3 & 7.9 & 5.1 & 0.6 & 0.4 & 0.0 \\
        Rebuffel & transformer & 4.7 & 5.5 & 5.0 & 1.7 & 0.5 & 0.1 \\ \hline
    \end{tabular}
    \caption{Breakdown of error types per-text, by system.  20 texts were included in the training corpus for each system.}
    \label{fig:all_system_errors}
\end{table*}

\begin{figure}[ht!]
    \resizebox{\columnwidth}{!}{%
    \begin{tikzpicture}
        \begin{axis}
            [
                ymin=0,
                every axis x label/.style={at={(current axis.right of origin)},anchor=south east},
                title={\textit{\textbf{Name errors}}},
                xlabel=Tenth of summary,
                ylabel=No. of Errors,
                x label style={at={(axis description cs:0.5,-0.1)},anchor=north},
                enlargelimits = false,
                xticklabels from table={name.dat}{X},xtick=data,
                legend style={at={(0.9,0.9)}},
            ]
            \addplot[maroon,dotted,thick] table [y=Wiseman,x=X]{name.dat};
            \addlegendentry{Wiseman}
            \addplot[purple,dashed,thick] table [y=Puduppully,x=X]{name.dat};
            \addlegendentry{Puduppully}
            \addplot[teal,thick] table [y= Rebuffel,x=X]{name.dat};
            \addlegendentry{Rebuffel}];
        \end{axis}
    \end{tikzpicture}
    }%
    \caption{Name errors in different tenths of the summary.}
    \label{Figure-Name-errors}
\end{figure}

\section{Submissions}

\subsection{Automatic approaches}
\subsubsection{Charles-UPF}

Charles University and Pompeu Fabra University submitted a system which detects errors using a three-stage process
\begin{enumerate}
    \item A rule-based NLG system is used to generate sentences with facts that can be derived from the game data.
    \item For each sentence in the NLG texts, a subset of the sentences in (1) is chosen based on semantic similarity to the target sentence.
    \item A language model is used to identify errors.  The input to the model is both the target sentence and the sentences in (2).  The model is trained on synthetic data as well as the training data.
\end{enumerate}
Note that the Charles-UPF system checks sentences separately, so it cannot detect errors that depend on document-level context, including Context errors and usage of \scare{\errword{only other}} (\autoref{sec:blindspot}).

\subsubsection{Eurecom}

The Eurecom system follows an approach inspired by earlier work on computational fact-checking \citep{DBLP:journals/pvldb/Karagiannis0PT20}. It focuses on identifying Number errrors, and also Word errors where the word maps in a straightforward way to the game data, such as errors in the usage of \scare{\errword{defeated}}.  A three-step process is used
\begin{enumerate}
    \item \textit{Claim identification:} Factual claims are extracted from the NLG text.
    \item \textit{Property identification}: The claims in (1) are expanded into full property specifications; for example the claim \textit{18 points} is expanded with the name of the player who is supposed to have scored these points.
    \item \textit{Claim verification}: The game data is queried using the property specifications in (2); incorrect claims are flagged.
\end{enumerate}

\subsubsection{National Institute of Japanese Literature}

The NIJL system  used different approaches for different types of errors:
\begin{itemize}
    \item \textit{Word and Name errors}:   A set of rules was used to identify Word and Name errors in the NLG texts.  These rules were tuned to the structure of game summaries, with different rule used for lead, middle, and end sections of the summaries.  The rules referred to the human reference texts as well as the game data.
    \item \textit{Number errors}:  A classifier was used to predict what relation each number represented.  A co-reference tool was used to resolve referring expressions such as \scare{\textit{he}}.
\end{itemize}

The NIJL system was the only submission which used the
human-written reference texts as well as game data when looking for accuracy errors; all other submissions just used the game data.

\subsection{Hybrid approaches}

\subsubsection{Laval University}

The Laval University approach was a hybrid system, which combined automatic analysis and human annotation.
\begin{enumerate}
    \item \textit{Pre-annotation}: a set of rules and classifiers are used to highlight potential accuracy errors in the NLG text. Row-column lookup on source data is used to identify potential Name and Number errors, and a multi-class, multi-label classifier is trained for Word, Context, and Not Checkable errors.
    \item \textit{Human annotation}: a single human annotator then annotated errors in the NLG text, using the pre-annotation of (1) to help them.
\end{enumerate}

The human annotation was much quicker than the protocol of \citet{INLG20accuracy}, because of the pre-annotation step.

We present two results for Laval University: a \scare{metric} result which is based purely on the results of the pre-annotation process, and a \scare{hybrid} result which is based on the full approach described above.

\section{Results}

The submissions were evaluated by computing their recall and precision against the gold-standard mistake list (GSML) which was based on the human annotated texts in the test set (\autoref{sec:data}).  In other words, for each submission, we calculated how many of the gold-standard mistakes were detected by that submission (recall), and how many of the mistakes detected by that submission were present in the gold-standard annotation (precision).  We calculated this at the level of both mistakes and tokens.

\autoref{fig:all_team_system_errors} shows the recall and precision of our submissions against the gold-standard manually annotated texts, for the 30 texts in the test set.  We can see that the Laval University hybrid approach  did best.  Amongst the automatic evaluations, the Charles-UPF system had the best recall and precision.

Tables \ref{fig:laval} to \ref{fig:njil} show recall/precision of the submissions for different types of mistakes, as well as overall.  We can see that the automatic techniques (Tables \ref{fig:charles-university} to \ref{fig:njil}) were unable to detect Context and Other errors, and only the Laval University (metric) system could detect Not Checkable errors (but at low precision and recall).  We can also see that none of the automatic systems did well at detecting Word errors; the best system, Charles-UPF, had around 50\% precision and recall.  Overall, this suggests that semantically more complex errors are harder to detect automatically, which is not surprising.

As a point of comparison, the Relation Generation metric \citep{wiseman-etal-2017-challenges}, which has been widely used by many previous papers to evaluate accuracy, can only detect Name and Number errors and has a recall of less than 40\% for these types of errors \citep{INLG20accuracy}.   This is considerably worse than the best-performing submissions to our shared task.

\begin{table}[ht]
    \resizebox{\columnwidth}{!}{%
        \begin{tabular}{|l|cc|cc|}
            \hline
            & \multicolumn{2}{c|}{Mistake} & \multicolumn{2}{c|}{Token} \\
			Team & recall & precision & recall & precision \\ \hline
            Laval University* & 0.841 & 0.879 & 0.668 & 0.859 \\
            Charles-UPF & 0.691 & 0.756 & 0.550 & 0.769 \\
            NIJL & 0.523 & 0.494 & 0.349 & 0.505 \\
            Laval University & 0.503 & 0.334 & 0.410 & 0.397 \\
            Eurecom & 0.080 & 0.311 & 0.046 & 0.202 \\ \hline
        \end{tabular}%
    }
    \caption{Results of the Accuracy Evaluation Shared Task for all submissions.  The * denotes the hybrid evaluation for Laval University.  All other submissions were metrics.}
    \label{fig:all_team_system_errors}
\end{table}

\begin{table}[ht]
	\resizebox{\columnwidth}{!}{%
		\begin{tabular}{|l|cc|cc|}
		    \hline
			& \multicolumn{2}{c|}{Mistake} & \multicolumn{2}{c|}{Token} \\
			Team & recall & precision & recall & precision \\ \hline
			Name & 0.920 & 0.938 & 0.929 & 0.919 \\
			Number & 0.862 & 0.881 & 0.832 & 0.854 \\
			Word & 0.679 & 0.731 & 0.561 & 0.685 \\
			Context & 0.750 & 0.400 & 0.733 & 0.367 \\
			Not checkable & 0.237 & 0.391 & 0.073 & 0.615 \\
			Other & 0.000 & - & 0.000 & - \\ \hdashline
			Overall & 0.841 & 0.879 & 0.668 & 0.859 \\ \hline
		\end{tabular}%
	}
	\caption{Laval University (hybrid evaluation) per-type results.}
	\label{fig:laval}
\end{table}

\begin{table}[ht]
	\resizebox{\columnwidth}{!}{%
		\begin{tabular}{|l|cc|cc|}
		    \hline
			& \multicolumn{2}{c|}{Mistake} & \multicolumn{2}{c|}{Token} \\
			Team & recall & precision & recall & precision \\ \hline
			Name & 0.750 & 0.846 & 0.759 & 0.862 \\
			Number & 0.777 & 0.750 & 0.759 & 0.752 \\
			Word & 0.514 & 0.483 & 0.465 & 0.529 \\
			Context & 0.000 & - & 0.000 & - \\
			Not checkable & 0.000 & - & 0.000 & - \\
			Other & 0.000 & - & 0.000 & - \\ \hdashline
			Overall & 0.691 & 0.756 & 0.550 & 0.769 \\ \hline
		\end{tabular}%
	}
	\caption{Charles-UPF (metric) per-type results.}
	\label{fig:charles-university}
\end{table}

\begin{table}[ht]
	\resizebox{\columnwidth}{!}{%
		\begin{tabular}{|l|cc|cc|}
		    \hline
			& \multicolumn{2}{c|}{Mistake} & \multicolumn{2}{c|}{Token} \\
			Team & recall & precision & recall & precision \\ \hline
			Name & 0.000 & - & 0.000 & - \\
			Number & 0.205 & 0.329 & 0.198 & 0.203 \\
			Word & 0.014 & 0.095 & 0.006 & 0.095 \\
			Context & 0.000 & - & 0.000 & - \\
			Not checkable & 0.000 & - & 0.000 & - \\
			Other & 0.000 & - & 0.000 & - \\ \hdashline
			Overall & 0.080 & 0.311 & 0.046 & 0.202 \\ \hline
		\end{tabular}%
	}
	\caption{Eurecom (metric) per-type results.}
	\label{fig:Eurecom}
\end{table}

\begin{table}[ht]
	\resizebox{\columnwidth}{!}{%
		\begin{tabular}{|l|cc|cc|}
		    \hline
			& \multicolumn{2}{c|}{Mistake} & \multicolumn{2}{c|}{Token} \\
			Team & recall & precision & recall & precision \\ \hline
			Name & 0.594 & 0.787 & 0.641 & 0.811 \\
			Number & 0.442 & 0.351 & 0.427 & 0.340 \\
			Word & 0.357 & 0.137 & 0.207 & 0.146 \\
			Context & 0.000 & - & 0.000 & - \\
			Not checkable & 0.500 & 0.190 & 0.200 & 0.407 \\
			Other & 0.000 & - & 0.000 & - \\ \hdashline
			Overall & 0.503 & 0.334 & 0.410 & 0.397 \\ \hline
		\end{tabular}%
	}
	\caption{Laval University (metric) per-type results.}
	\label{fig:laval_metric}
\end{table}

\begin{table}[ht]
	\resizebox{\columnwidth}{!}{%
		\begin{tabular}{|l|cc|cc|}
		    \hline
			& \multicolumn{2}{c|}{Mistake} & \multicolumn{2}{c|}{Token} \\
			Team & recall & precision & recall & precision \\ \hline
			Name & 0.358 & 0.974 & 0.258 & 0.974 \\
			Number & 0.696 & 0.419 & 0.672 & 0.419 \\
			Word & 0.350 & 0.301 & 0.245 & 0.310 \\
			Context & 0.000 & - & 0.000 & - \\
			Not checkable & 0.000 & - & 0.000 & - \\
			Other & 0.000 & - & 0.000 & - \\ \hdashline
			Overall & 0.523 & 0.494 & 0.349 & 0.505 \\ \hline
		\end{tabular}%
	}
	\caption{National Institute of Japanese Literature (metric) per-type results.}
	\label{fig:njil}
\end{table}

\subsection{Error analysis: The blind spot of metric submissions}
\label{sec:blindspot}
To explore our intuition that complex errors were harder for the automatic systems to find, we performed a preliminary error analysis on the 84 mistakes (of 622) that were missed by all automatic submissions (the blind spot).  We categorised each mistake based on the type of sentence that contained it:

\paragraph{Simple:}Only 27 of the mistakes were simple, such as an incorrect attribute for an entity, or an incorrect name for a set of attributes.  An example is `\textit{\errname{Buddy Hield} led the second unit with a season - high 29 points , along with one assist , one rebound and one steal}', where the statistics belonged to Eric Gordon.

\paragraph{Comparison:}26 of the mistakes involved the comparison of how two teams fared in a quarter/half, or how their statistics compared in the game.  An examples is `\textit{The Nets got off to a \errword{quick start} in this one, \errword{out-scoring} the Kings 28-28 right away in the first quarter.}', where the tie of 28 points in the first quarter is incorrectly described.  Many of these mistakes also involved getting the X-Y numbers wrong.

\paragraph{Only other:}14 of the mistakes were in clauses like `\textit{The \errword{only other} Net to reach double figures in points was Ben McLemore}`.  This requires models and metrics to determine:
    \begin{itemize}
        \item That Ben McLemore had double-figures and was a Nets player.
        \item Which other Nets had double-figures.
        \item That all such players have been mentioned previously.
    \end{itemize}
\paragraph{Data outwith game:} 11 of the mistakes required data from outwith the game being summarised, including averages over prior games (8 mistakes), and the upcoming game schedule (3 mistakes).

\paragraph{Player groups:} 6 mistakes incorrectly described a group of players, such as a duo.\\

45\% of blind spot mistakes involved Word, Context, and Not-Checkable errors, compared to only 30\% overall in the GSML.  In addition, only 8\% of blind spot mistakes were cardinal numbers, despite these constituting 33\% of the GSML.  It is important that we do not miss blind spot mistakes as whilst they are only 14\% of the current GSML, this proportion could increase as systems become better at avoiding simple errors.

\section{Conclusion}
Neural data-to-text systems need to be able to produce accurate texts in order to be genuinely useful in most NLG use cases.  An essential prerequisite to improving accuracy is being able to measure and evaluate accuracy.

We believe that the evaluation techniques submitted to our shared task represent a major advance in the state of the art.  We encourage participants and others to continue developing better-performing techniques for this key evaluation task.

\section*{Acknowledgments}
We are very grateful to all of the participants in the shared task, for their hard work in exploring very diverse approaches to the problem of finding accuracy errors! Several other teams were unable to complete a submission by our deadline; they had very interesting ideas and we encourage them to continue working on their ideas.  We are also very grateful to Samira Shaikh and the members of the Aberdeen CLAN group for their help and advice.   Last but not least, we are very grateful for the hard work of our Mechanical Turk annotators, in creating our training and test data.  Craig Thomson's work is supported under an EPSRC NPIF studentship grant (EP/R512412/1).

\bibliographystyle{acl_natbib}
\bibliography{accuracySharedTask}

\end{document}